# Supervised Learning Achieves Human-Level Performance in MOBA Games: A Case Study of Honor of Kings

Deheng Ye, Guibin Chen, Peilin Zhao, Fuhao Qiu, Bo Yuan, Wen Zhang, Sheng Chen, Mingfei Sun, Xiaoqian Li, Siqin Li, Jing Liang, Zhenjie Lian, Bei Shi, Liang Wang, Tengfei Shi, Qiang Fu, Wei Yang, and Lanxiao Huang

*Abstract*—We present JueWu-SL, the first supervised-learning-based artificial intelligence (AI) program that achieves human-level performance in playing multiplayer online battle arena (MOBA) games. Unlike prior attempts, we integrate the macro-strategy and the micromanagement of MOBA-game-playing into neural networks in a supervised and end-to-end manner. Tested on Honor of Kings, the most popular MOBA at present, our AI performs competitively at the level of High King players in standard 5v5 games.

## I. Introduction

MULTIPLAYER online battle arena (MOBA) games, e.g., Dota, Honor of Kings, and League of Legends, have been considered as an important and suitable testbed for artificial intelligence (AI) research due to their considerable complexity and varied playing mechanics [1]–[4]. The standard game mode of MOBA is 5v5, where two opposing teams of five players each compete against each other. In this mode, each individual in a team has to control the actions of one hero in real time based on both the situation dynamics and the team strategy. During the game, a hero can grow stronger by killing enemy heroes, pushing turrets, killing creeps and monsters, and so on. The goal for players in a team is to destroy their enemy's main structure while protecting their own. In MOBA, the gameplay is also varied as it involves two factors: macro-strategy, i.e., "where to go" on the game map for a hero, and micromanagement, i.e., "what to do" when the hero reaches the expected location. Despite such complexity and mechanism, the methodological explorations on AI systems for MOBA-game-playing are still very limited [5].

In this article, we explore the potential of supervised learning (SL) for building MOBA Game AI agents. We are motivated by the following practices and observations. First, SL is usually the first step to build AI agents and plays a significant role in achieving human-level performance for many games. For example, the SL network is reused as the policy network of reinforcement learning (RL) for the Game AI of Go [6]; a competitive SL model is incorporated as human knowledge to facilitate the policy exploration of AI agent in general real-time strategy (RTS) games, such as StarCraft [7]; and pure SL can also produce human-level AI agents in complex video games, such as Super Smash Bros [8]. Second, recent studies on policy distillation [9], [10] have empirically shown that, with effective knowledge representation, SL can generalize to sequential prediction tasks. In MOBA games, as experienced human players often judge the game situation before initiating any acts, the actions (labels) taken in effect should naturally contain sequential information if properly extracted. In this sense, we could "distill" the game knowledge of human experts in a supervised manner. Specifically, suppose that we could have well-annotated labels, in which strategies and actions of a team are embedded, and expressive representations of every mini combat at each timestep in a MOBA game, SL would potentially "distill" an effective mapping from mini combats to labels given sufficient training data.

However, such distillation-like mapping is very challenging due to the modeling of both the macro-strategy and the micromanagement of MOBA players. The macro-strategy, as an abstract concept in MOBA Esports, has no explicit and mathematical definitions per se. The micromanagement is also a decision-making process of vast complexity due to the hero's complicated atomic actions and its dynamic game environment (the obstacles, allies, enemy heroes, and turrets are evolving and ever-changing over time).

To this end, we propose an SL method to build MOBA Game AI agents. We design multiview intent labels to model

MOBA player's macro-strategy; we design hierarchical action labels for modeling micromanagement. Based on the label design, we formulate the action-taken process of a MOBA Game AI agent as a hierarchical multiclass classification problem. We resort to deep convolutional and fully connected (FC) neural networks with multimodal features to solve the problem. Unlike previous attempts, our method simultaneously learns the AI agent's macro-strategy and micromanagement in a supervised and end-to-end manner. Here, end-to-end refers to the combination of macro-strategy and micromanagement modeling, rather than the whole training process. Specifically, the output of the macro-strategy component is embedded and then provided to the micromanagement component as an auxiliary task, which has also been applied in other problem settings [11]–[13]. Furthermore, we develop a scene-based sampling method for training data preprocessing.

Based on our method, an AI agent named JueWu-SL (JueWu means "Insight and Intelligence" in Chinese, and SL refers to supervised learning) is trained using the historical data of Honor of Kings, whose 1v1 and 5v5 modes have been used extensively as testbeds for recent Game AI advances [2], [3], [14]–[16]. JueWu-SL is tested by competing against some existing AI methods and human players. Encouraging results demonstrate that our method can produce extremely competitive AI agents with top human-level performance. Such results justify and indicate the great potential of meticulously designed SL systems for MOBA Game AI. We expect our findings to encourage further research in developing MOBA Game AI programs using SL.

To sum up, the contributions of this article are as follows.

1) We introduce an SL method for Game AI in MOBA, which allows efficient incorporation of game knowledge from human experts, leading to the fast manufacturing of high-level AI agents. To the best of our knowledge, this is the first work for Game AI in MOBA that simultaneously models micromanagement and macro-strategy, leveraging convolutional and FC neural networks with multimodal features. We further propose a scene-based sampling method to divide a whole MOBA gameplay into easily tuned game snippets, making it possible to polish the AI performance in varying game situations.

2) Extensive experiments show that our AI performs competitively in matches against High King[1] level human players in Honor of Kings. This is the first time for SL-based AI in MOBA games to achieve such performance. Qualitative analysis is further provided to understand why the AI works in practice.

The remainder of this article is organized as follows. We review related work in Section II. We then briefly introduce some background knowledge of MOBA games, before we present our Game AI method in Section IV. Section V covers experimental results and discussions. We conclude in Section VI.

## II. Related Work

Our work applies neural networks to system-level AI development for playing games. We, thus, summarize representative studies on system-level AI for different types of games, covering Non-RTS, RTS, and MOBA games. Partial Game AI solutions, e.g., predicting match results or time to die for heroes [17], [18], will not be discussed in detail.

### A. Non-RTS Games

The early method for Game AI was search-based. One classic work was DeepBlue that defeated the world chess champion [19]. Follow-up studies then focused on designing AI agents for more complex games. For example, AlphaGo, a Go-playing agent that defeated top professionals, was developed via tree search and neural network [6] and was later enhanced to AlphaZero [20]. Many other types of Go agents were also proposed, e.g., FineArt (the champion of the UEC Cup in 2017) from Tencent, and OpenGo (open-sourced reimplementation of AlphaGo) from Facebook [21]. In the meantime, RL-based AI agents were also developed for Atari series [22] and first-person shooter (FPS) games [23], [24]. The great success of AI for these game types has inspired research for complex strategy video games, such as general RTS games and MOBA games.

### B. General RTS Games

During the past several years, StarCraft has been used as the testbed for Game AI research in RTS. Methods adopted by existing studies include rule-based, supervised, reinforcement, and their combinations [25], [26]. For rule-based approach, a representative is SAIDA, the champion of StarCraft AI Competition 2018.[2] Another recent AI program AlphaStar combines SL and RL and achieves grandmaster level in playing StarCraft 2 [7].

### C. MOBA Games

As a subgenre of RTS, MOBA games emphasize complex control and team collaboration and become a suitable testbed for AI research. However, we see limited attempts on system-level AI development for playing MOBA games. State-of-the-art studies for MOBA 5v5 games are from Tencent and OpenAI, as summarized in Table I. Tencent HMS, a macro-strategy model for Honor of Kings [2], was developed by dividing the whole game into different phases and then training a model with user data for each phase in a supervised manner. However, HMS only modeled macro-strategy, i.e., predicting where to go in the game map, and overlooked micromanagement, i.e., specific action to take. Thus, HMS cannot be solely regarded as a system-level AI solution for MOBA games. Furthermore, the macro-strategy in JueWu-SL differs from HMS in: 1) introducing multiview global and local intents to encode player's macro-strategy in MOBA games; 2) eliminating explicit cross-communication and phase attention used in the HMS model; and 3) unified feature representations for both macro-strategy and micromanagement. Another study was recently proposed by OpenAI, which introduced an AI for playing Dota 2 5v5 games (Dota 2 is a popular MOBA game), called OpenAI Five, with the ability to defeat professional human players. OpenAI Five is based on deep RL via self-play and is trained using a scaled-up proximal policy optimization (PPO) [27]. The implementation details of OpenAI Five are open since December 13, 2019 [4]. In contrast, we contribute an SL approach and our attempt sheds light on the potential of supervised neural models in building AI for the multiagent MOBA 5v5 games.

---

[1]In Honor of Kings, a player's game level (or rank) can be No-rank, Bronze, Silver, Gold, Platinum, Diamond, Heavenly, King (also known as Champion), and High King (the highest level), in ascending order. A King player's rank is further distinguished by the number of stars gained. The more stars the higher rank. King players reaching 50 stars are promoted to High King.

[2]See https://github.com/TeamSAIDA/SAIDA

TABLE I
COMPARING AI LEARNING SYSTEMS FOR MOBA GAMES

| Work | Game | Method | Notes |
|---|---|---|---|
| OpenAI Five | Dota 2 | Reinforcement | Details released in Dec. 2019 |
| Tencent HMS | Honor of Kings | Supervised | Only macro-strategy studied |
| This work | Honor of Kings | Supervised | End-to-end, complete AI |

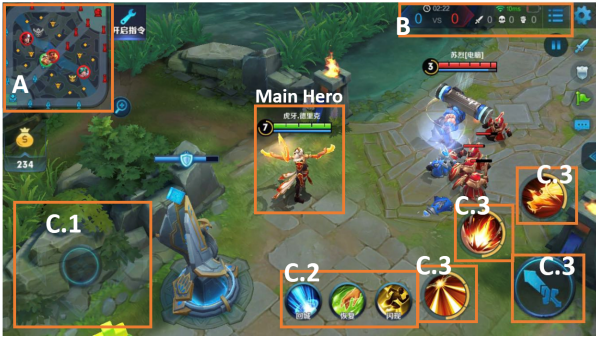

Fig. 1. Game UI of Honor of Kings. The hero controlled by the player is called the " main hero." Bottom left is the movement controller (C.1), while the right-bottom set of buttons are ability controllers (C.2 and C.3). Players can observe game situation via the screen (local view) and obtain game states with the dashboard (B) and obtain global view with top-left minimap (A).

## III. BACKGROUND

Fig. 1 shows a typical game scenario in Honor of Kings. The player controls the main hero, located at the screen center, with a fine-grained local view of the environment, which includes observable characters' position, health, and so on. In contrast, the minimap on the top left gives a coarse-grained view of the global map and the whole game situation, which includes the information of turrets, creeps, and the observable heroes. In terms of hero control, the player uses the joystick (at the bottom left) and buttons (at bottom right) to move the main hero and trigger its skills and normal attacks. The buttons at the bottom middle are used to trigger the assistive skills, such as Return Base and Summon Skill. With this interface, the micromanagement of a player, i.e., "what to do concretely" given the game state, is all the controls for the main hero, including movements, normal attack, and skill triggering, while the macro-strategy refers to the player's intention on "where to go on the map" and "what to do globally" during a certain period of game time, in order to gain rewards effectively.

Given a certain game scene in MOBA with observable game states and hero information, an experienced MOBA player is able to judge what actions to take and where to go on the map. For example, for the case in Fig. 1, a high-level player may directly attack the enemy hero using skills, given the fact that the main hero is in a safe position and its level is higher than the enemy's (level 7 versus level 3). The actions taken here inherently and naturally have a correspondence to the game states.

## IV. METHOD

This section describes our method in detail, including label design, features, model, and data sampling.

### A. Overview

We aim to build an AI model for playing MOBA games via SL. To this end, we design features and labels suitable for MOBA games by drawing inspirations from professional MOBA players. The features have two parts: vector feature, which represents the game states, and image-like feature, which represents the maps. The labels also have two parts: intent labels for macro-strategy and action labels for micromanagement. Specifically, the intent labels, abstracted from player's attention in a game, are multiviewed, with global and local intents. The action label has a hierarchical multiclass structure, which contains two parts. The first part is the action to take at a high level, such as move; the second one is how to take a specific action, such as the discretized direction to move. Note that the action labels are the output labels of the model, while the intent labels are used as an auxiliary task during training to improve the model performance. With such feature and label design, the AI modeling task can be solved as a hierarchical multiclass classification problem. Specifically, a deep neural network model is trained to predict what actions to take for a given game situation. For online deployment, the model predicts the output labels hierarchically in two steps: 1) the high-level action to take and 2) the execution specifications of that action.

In the following, we use the popular MOBA game Honor of Kings as an example to illustrate how the proposed method works. Note that the methodologies proposed in this article are general to MOBA games since different MOBA games share similar playing mechanism. For example, our hierarchical action design and multiview intent design can be easily adapted to other MOBA games, such as Dota and League of Legends.

### B. Multiview Intents for Macro-strategy

Similar to how players make decisions according to the game map, macro-strategy predicts regions for an agent to move to. Based on human knowledge, regions where attack takes place can be an indicator of players' destination because, otherwise, players would not have spent time on such spots.

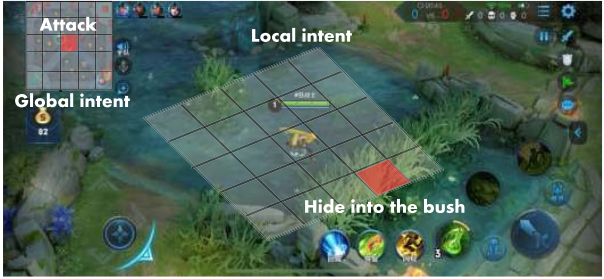

Fig. 2. Multiview intent label design for abstracting player's macro-strategy.

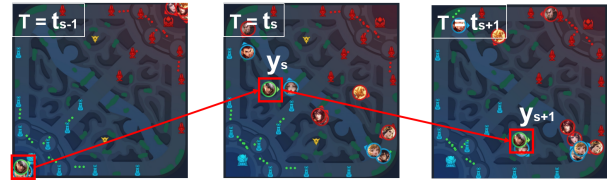

Fig. 3. Example of the global intent label extraction.

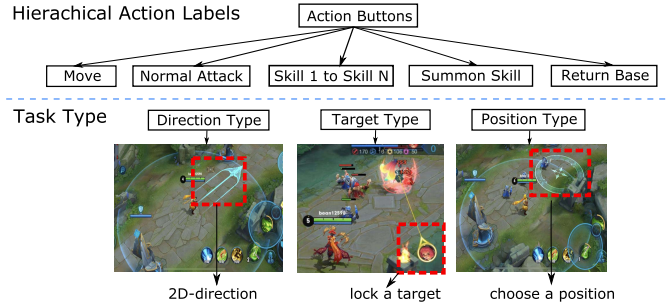

Fig. 4. Hierarchical action label design for representing micromanagement in MOBA.

We, thus, define ground-truth regions as the regions where players conduct their next attack. However, an occasional attacking behavior in a given spot does not have to be the goal of a player, and it might be a consequence of moving somewhere else and encountering an opponent on the way. We, thus, only consider regions where continuous attacking behaviors exist. Based on such consideration, we design multiview intent labels, including a global intent and a local intent label to model macro-strategy. Fig. 2 illustrates the idea of multiview intent labels.

The global intent of the player is the next target or region, where the players may attack creeps to gain gold, attack turrets for experience, and others. It can be identified from the game situation, which is very important for RTS. For example, at the beginning of the game, the jungle hero normally focuses on the jungle region, where he can kill the monsters to grow the game level faster. To define the global intent label, we divide the square MOBA map into course-grained $N \times N$ regions (for Honor of Kings, $N = 24$, i.e., a total of 576 intent classes is defined for this game). Then, the global intent label is defined as the numbered region on the map where the next series of attacks happen.

Local intent is short-term planning in local combat, including gathering in the bush or safe points, retreating to the nearby turret, or waiting for target heroes, before finally attacking a target. To define local intent label, we divide the local-view map into fine-grained $M \times M$ regions (for Honor of Kings, $M = 12$), where the resolution of each local region is designed to be roughly equal to the physical edge length of a hero. Then, the local intent label is one of the numbered regions from the local-view map of the agent, which is extracted from the halfway position of the players between two attack events. Taking Honor of Kings for example, we have analyzed that the interval between two attack events is about 6.5 s on average. This local intent label provides additional intermediate information for learning macro-strategy.

Fig. 3 gives a concrete example of the global intent extraction. Let $s$ be one session in a game that contains several frames, and $s - 1$ indicates the previous session of $s$. Let $t_s$ be the starting frame of $s$. Note that a session ends along with attack behavior; therefore, there exists a region $y_s$ in $t_s$ where the hero conducts attack. As shown, the label for $s - 1$ is $y_s$, while the label for $s$ is $y_{s+1}$. Intuitively, by setting up labels in this way, we expect agents to learn to move to $y_s$ at the beginning of the game. Similarly, agents are supposed to move to appropriate regions given the game situation.

Finally, both of these intent labels are trained with the multimodal features, and their softmax results are then fed as auxiliary features for the task of action labels training, which will be shown in Fig. 7.

### C. Hierarchical Action Design for Micromanagement

We design hierarchical action labels to control a hero's micromanagement, as illustrated in Fig. 4. Specifically, each label consists of two components (or sublabels), which represent level-1 and level-2 actions. The first one, i.e., level-1 action, indicates which action to take, including move, normal attack, skill 1, skill 2, skill 3, summons, return base, and so on. The second one, level-2 action, tells how to execute the action concretely according to its action type. For example, if the first component is the move-action, then the second one will be a 2-D-direction to move along. When the first component is normal-attack, the second one will be a target to attack. If the first component is skill 1 (or 2, or 3), the second one will depend on the action type of level-1 action because all the skills 1–3 of different heroes belong to three types: direction type, target type, and position type. For example, if it is the position type, the second one will be a position where the action will take effect.

### D. Multimodal Features

Based on the game expert's experience, we develop a comprehensive list of features, including vector feature and image-like feature, to represent the game situations and combats during a MOBA game.

In Fig. 5, we illustrate the extraction process of vector and image-like features. The vector feature is made up of observable hero attributes and game states. Hero attributes consist of current health point (hp), historical hp in past frames, skill cool down, damage attributes, defense attributes, gold, level, buff, historical coordinates in past frames, and so on. Game states are composed of team kill difference, gold difference, game time, turrets difference, and so on. For convolution, we use image-like features extracted from the game engine, rather than the image. We develop local image-like and global image-like features that are extracted from the player hero's local view map and the global minimap, respectively. Fig. 6 gives an example of a local image-like feature for the damage channel that describes one skill effect from the enemy hero. On the other hand, for global image-like features, the channels are like matrices of the positions of observable heroes, soldiers, monsters, and other information. These image-like features represent high-level concepts that are used to help reduce the number of convolution layers required. Feature setups will be covered in Section V-A.

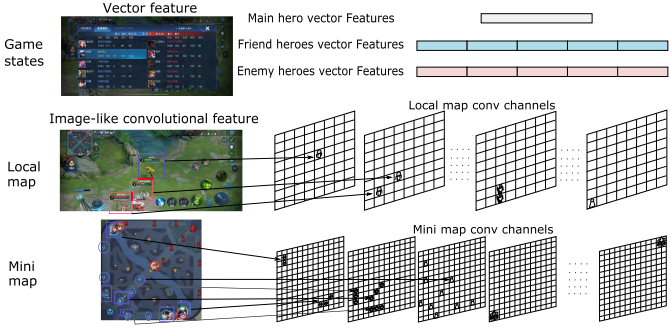

Fig. 5. Illustrating multimodal feature extraction. We extract observable game states and game unit attributes from the dashboard, as vector features. We extract image-like features from the local and minimap of the main hero.

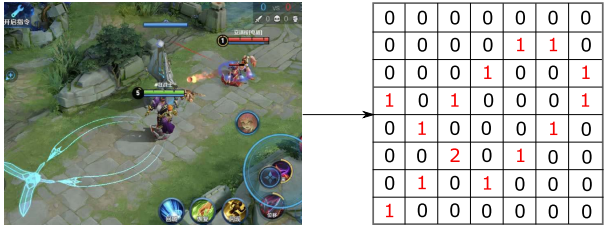

Fig. 6. Illustration of local image-like feature extraction. Transforming a local view to a local image-like feature channel: damage channel. In the local view, an enemy is releasing a skill to attack the main hero. The attacked range is shown as two fluorescent curves. This local view is transformed into the image-like feature channel, for which the corresponding positions of the red curves are assigned with values according to the real-time damage.

### E. Model

To train a model for a hero, a training dataset of $\{(x_i, y_i) | i = 1, \ldots, n\}$ is extracted from historical games. Each instance $(x_i, y_i)$ is extracted from the current and past frames, where $x_i$ and $y_i$ correspond to the multimodal features and the intent and action labels of the player, respectively. Concretely, $x_i = \{x_i^v, x_i^g, x_i^l\}$, where $x_i^v, x_i^g$, and $x_i^l$ are the vector feature, global image-like feature, and local image-like feature, respectively. $y_i = (y_{ai}^0, y_{ai}^1, y_{bi}^g, y_{bi}^l)$, where $y_{ai}^0 \in \{1, \ldots, m\}$ and $y_{ai}^1$ are the level-1 and level-2 action label, and $m$ denotes the number of level-1 actions; $y_{bi}^g$ and $y_{bi}^l$ are the global and local intent labels, respectively. Based on the data, a deep neural network model is proposed for Honor of Kings in Fig. 7, which is also suitable for any MOBA games. In this model, the global and local image-like features $x^g$ and $x^l$ are mapped to $h_g(x^g; \psi_g)$ and $h_l(x^l; \psi_l)$, respectively, by deep convolutional neural networks, where $\psi_g$ and $\psi_l$ are network parameters. Then, they are concatenated to predict the global and local intent labels as

$$p^{m+1} = e_g\big([h_l(x^l; \psi_l), h_g(x^g; \psi_g)], \phi_g\big)$$
$$p^{m+2} = e_l\big([h_l(x^l; \psi_l), h_g(x^g; \psi_g)], \phi_l\big)$$

where $\phi_g$ and $\phi_l$ are the network parameters. In addition, the vector feature $x^v$ is mapped to $h_v(x^v; \psi_v)$, where $\psi_v$ is a network parameter. For vector features, we perform feature grouping to split them into 11 parts, as shown in Fig. 5. We then feed each of them through FC layers before they are concatenated. Finally, the model concatenates all the five above vectors and maps it to the final hidden layer, as

$$h(x) = h_m\Big([h_v(x^v; \psi_v), h_g(x^g; \psi_g),$$
$$h_l(x^l; \psi_l), p^{m+1}, p^{m+2}]; \mu\Big)$$

where $\mu$ denotes parameters of FC. Then, $m + 1$ different functions $\{f^i(\cdot; \omega_i) | i = 0, \ldots, m\}$ are utilized to make a prediction $p = (p^0, \ldots, p^{(m+2)})$, where $p^i = f^i(h(x); \omega_i)$ for $i = 0, \ldots, m$. To fit a prediction to preprocessed human data, a multitask loss is proposed as a weighted sum of losses

$$\ell(p, y) = w_{a0}\ell_{CE}(p^0, y_a^0) + w_{a1}\ell_{CE}(p^{y_a^0}, y_a^1)$$
$$+ w_{bg}\ell_{CE}(p^{m+1}, y_b^g) + w_{bl}\ell_{CE}(p^{m+2}, y_b^l)$$

where the first two items are for action labels, while the latter two are for intent labels, and $\ell_{CE}$ denotes the cross entropy loss, as we discretize intent and action labels. For action labels, we only sum two losses because the label $y$ does not provide any information for the actions: $\{y \neq y_a^0 | y = 1, \ldots, m\}$.

To train such a model, the objective is introduced as follows:

$$\min_{\theta} \sum_{i=1}^{n} \ell(p_i, y_i) + \lambda \|\theta\|_2^2$$

where $\theta = \{\psi_v, \psi_l, \psi_g, \phi_l, \phi_g, \mu, \omega_0, \omega_1, \ldots, \omega_m\}$ and $\lambda$ is the regularization parameter for avoiding overfitting. This objective can be solved using various stochastic gradient methods, such as Adam [28].

### F. Data Preprocessing

While the model is important for our AI, it cannot perform well without in-depth data preprocessing.

*1) Scene Identification:* As a preprocessing step, a MOBA game is segmented into different scenes because the action trajectories of a hero vary to achieve different purposes. Based on trajectory analysis, we define the following game scenes for MOBA (index (1–6) denotes priority).

1) *Push-Turret:* Attack enemy turret to occupy vision and control of the battlefield.
2) *Combat:* Encounter enemy hero and initiate a fight.
3) *Lane-Farm:* Attack waves of minions to gain gold and experience in the lane area.
4) *Jungle-Farm:* Attack monsters to gain gold and experience in the jungle area.

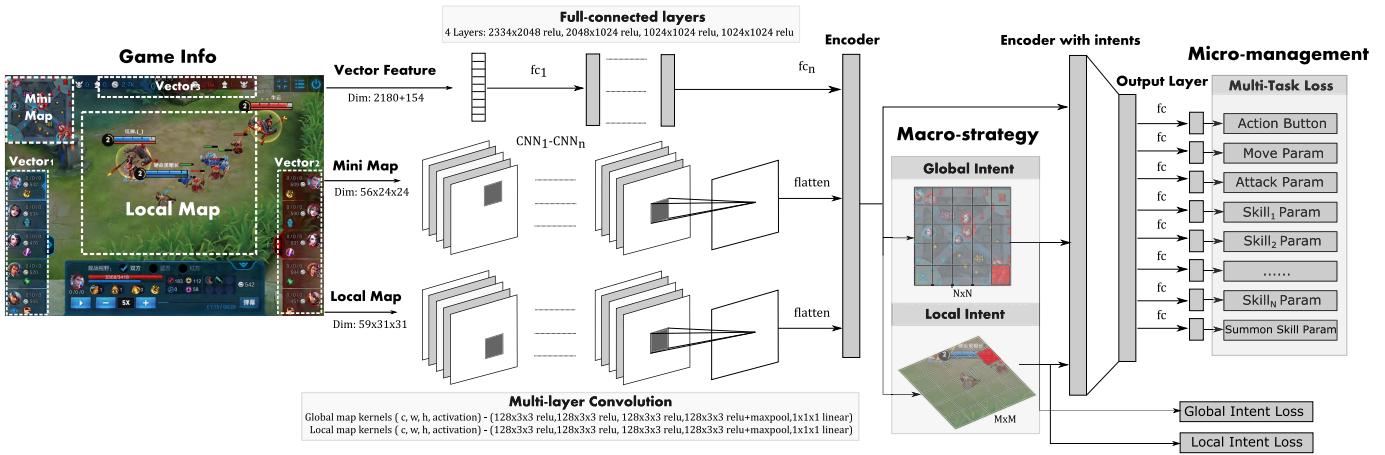

Fig. 7. Our multimodal and multitask deep neural network. The main structure consists of two parts: multimodal encoders and multitask output heads. Encoders: vector encoder with multilayer perceptrons and image encoder with multilayer convolution. Output heads: auxiliary macro-strategy tasks and micromanagement tasks.

5) *Return:* Return home base for recovering hero state or defending crystal.
6) *Navigation:* Move to another region in the map for macro-strategy intents.

Most of the scenes are strongly related to the attacking targets and distances between game units, i.e., target-oriented. For example, the Push-Turret Scene can be identified from the trajectory of attacking an enemy turret (a hero has to be in the attack range of the enemy turret in order to attack it). The Return Scene is related to a session that a hero returns to the home base for status recovery (there has to be no interruption from enemies over a period of frames for a successful return). The Combat Scene is identified by checking whether heroes are in a continuous fight. After identifying these target-oriented scenes, the rest of the scenes are categorized as the Navigation Scene.

*2) Data Tuning Under Each Scene:* Such a segmentation keeps the playing style of a given hero similar within each scene, which makes it convenient to tune the hero's performance in each one. Taking Navigation for example, one mainly wants to take the move action rather than releasing skills aimlessly. Therefore, the Move action should account for a large proportion of all actions in Navigation. However, in Combat, the focus should be attacking and skill releasing, together with move actions to avoid being attacked. Taking the DiaoChan hero in Honor of Kings as an example, empirically, the proportion of its move, normal attack, skill 1, skill 2, and skill 3 are tuned to 3:1:1:1:1 in Combat and 10:1:1:1:1 in Navigation. Note that these scenes sometimes can be overlapped. For example, when the opponent heroes stay under the turret, the pushing turret scene is also a combat scene. Once overlapped, the scenes categorization follows the abovementioned priority (1–6).

*3) Data Tuning Across Scenes:* After segmentation, the data are imbalanced between different scenes. The reason for this imbalance is the intrinsic design of the MOBA game. For example, turret pushing happens less frequently than combat. The jungle resources, namely, the monsters, are refreshed every few minutes, which limits the proportion of this scene.

Besides, the distributions of these scenes are different in terms of the roles of the hero. For example, because the role of a support hero is to use move to detect opponents and support teammates, most of its game scenes will fall into Navigation. However, the performances of AI in those scenes are equally important. Thus, we uniformly balance these scene data by downsampling to enable AI to be adequately trained in every scene. The downsampling ratio of these scenes is tuned hero by hero. Note that scenes are segmented only for off-line training. There is no difference in training between scenes. For online use, AI directly predicts actions based on the output of the model shown in Fig. 7.

*4) Move Sample Enhancement:* Move is important for all MOBA scenes. To compute Move actions effectively, we further amend the move directions for all game scenes. To expand, the move direction of a player is often random or useless, so that if the label of move direction is only computed using the single frame, it will be noisy. We thereby define the label of move direction as the difference between the position of the current frame and the position of a future frame, i.e., after $N$ frames, depending on whether the hero is in combats. In the Combat Scene, we can set $N$ as a fine-grained step since high-level players treat every move action seriously. In other scenes, it can be a coarse-grained step, since the player usually executes meaningless move actions. For the case of Honor of Kings, we use $N = 5$ (0.33 s) for the Combat Scene and $N = 15$ (1 s) for others.

*5) Attack Sample Normalization:* Another important sampling is target selection for the Combat and Push-Turret scenes. In the raw dataset, the examples of target selection for attacking are imbalanced between low-damage high-health (LDHH) heroes and high-damage low-health (HDLH) heroes. There are many more examples of attacking an LDHH hero than those of attacking an HDLH hero due to their different health property. Without downsampling, the model will prefer to attack the LDHH hero. However, the prior target normally is an HDLH hero, which is the key to win a local battle. To address this issue, we propose a hero-attack sampling method, termed attack sample normalization, which samples the same number of examples for one whole attacking process for all heroes. Experiments in real games indicate that attack sample normalization dramatically improves the quality of target selection.

## V. Experiments

### A. Experimental Setup

*1) Dataset:* The raw dataset comes from games played by the top 1% human players in Honor of Kings, from which we extract training samples. Each sample contains features, labels, label weights, frame numbers, and so on. We specify the number of parameters, such as the hero's ID, game level, camp, performance score, and date of the gameplay, for data filtering. There is an overall performance score of each player after the game and a real-time score of each player during the game based on the player's performance. We filter out the poorly performed games of top players with a performance score threshold, which is empirically set as the overall score that exceeds 90% of players using the same hero. For example, for the DiaoChan hero, such a score is set to no less than 10.0. We then preprocess, shuffle, and store the data in the HDF5 format. After preprocessing, only one-twentieth of the frames will be kept on average. Finally, 100 million samples from about 0.12 million games are extracted for one hero in our experiments. Each dataset of a hero is randomly split into two subsets: 1) a training set of around 90 million samples and 2) a test set of about 10 million samples.

*2) Model Setup:* Recall the three types of features, i.e., vector, local view, and global view. The vector feature has 2334 elements, of which 2180 are for the ten heroes and the remaining 154 are for the player's hero. The local view of a hero is set as a square with the hero at the center, of which the edge length is 30 000. It is then divided into 31*31 grids, to keep the grid length close to hero length, since the hero length is around 1000. Based on this resolution, the local image-like feature is of shape 59*31*31, i.e., it has 59 channels. The global map is divided into 24*24 grids (the edge length of the global map is 113 000), and the global image-like feature is of shape 56*24*24, i.e., it has 56 channels.

We train one model for one AI hero. For each hero, we use 16 Nvidia P40 GPUs to train for around 36 h. We use Adam [28] with initial learning rate 0.0001. The batch size is set to 256 for each GPU card. The weights of the four losses, i.e., $w_{a0}$, $w_{a1}$, $w_{bg}$, and $w_{bl}$, are of the same scale with our training set and are, thereby, tuned to 1 empirically. $\lambda$ for regularization is set to 1. In Honor of Kings, one game frame is about 66.7 ms (15 frames equal to 1 s). When in usage, the total reaction time of AI is 188 ms, including the observation delay (OD), which is set to 133 ms (two frames), and the reaction delay (RD), which consumes about 55 ms and is made up of processing time of features, model, results, and the transmission. Note that the OD can only be set as an integer.

*3) Metrics:* The most convincing way to evaluate a Game AI program is to deploy it online to test. We, thereby, deploy the AI model into Honor of Kings. Similar to the evaluations of AlphaGo [6], we evaluate the following aspects.

TABLE II
Comparison With Baseline Methods

| Tournament | VS BT | VS HMS | VS Two-stage |
|---|---|---|---|
| Win rate [CI] | 1.0 [0.96; 1] | 0.86 [0.78; 0.92] | 0.92 [0.85; 0.96] |
| Team kill | 19.3 : 5.5 | 26.4 : 17.5 | 29.8 : 17.0 |
| Game duration | 12.1 mins | 16.0 mins | 14.5 mins |
| Gold per min | 2.6k : 2.0k | 2.9k : 2.6k | 2.8k : 2.5k |
| Turret | 5.6 : 2.1 | 6.8 : 5.0 | 6.5 : 4.2 |
| Overlord&Tyrant | 1.5 : 0.6 | 2.9 : 1.9 | 2.4 : 1.6 |

1) Comparison against baseline methods. These methods include: 1) behavior-tree (BT), which is a classic search-based method for Game AI and has been used in various games as the internal AI, including StarCraft, Dota, and Honor of Kings; 2) HMS: as mentioned earlier, because HMS [2] cannot be used as a complete AI solution alone, a direct comparison is not possible, and we, thus, embed the output of HMS to our micromodel; and 3) two-stage: the macro-strategy and micromanagement models are trained separately and used in two stages. Specifically, the macromodel first predicts where to go in the map, and the micromodel predicts what to do after reaching the destination.
2) The performance of AI playing against human teams. We conduct matches in standard MOBA 5v5 mode, i.e., five AI agents play against five human players. This examines JueWu-SL's combat ability as a team.
3) The performance of AI playing with human teammates. We use AI + human team to play against the human team. This examines JuwWu-SL's collaboration with the human.

### B. Experimental Results

*1) Comparison With Baseline Methods:* We run a tournament among JueWu-SL and other AI programs in Honor of Kings. We run 100 matches for each comparison. The results are summarized in Table II. JueWu-SL outperforms BT, HMS, and Two-Stage with 100%, 86%, and 92% win rate, respectively. We also compute the exact binomial confidence interval (CI) [29] for each win rate. We can see that our AI performs stronger than baseline methods in real games, in terms of team kill, team gold, turrets destroyed, and so on.

*2) Playing Strength Against Human:* As mentioned earlier, a player's rank in Honor of Kings, from low to high, can be No-rank, Bronze, Silver, Gold, Platinum, Diamond, Heavenly, King, and High King (the highest level; 50-star King). A King player's rank is further distinguished by the number of stars gained. King players reaching 80 stars are very rare. We invited players of 80-star King level to play against JueWu-SL. All invited members were top-ranked players in the Sichuan Province of China. They often played as a team and can collaborate well. To thoroughly examine our AI's ability against the human team, the human team was encouraged to try different game strategies. Note that before we invited such top-notch human players, we had extensively tested JueWu-SL to make sure that it can easily defeat entry-level King players (top 1% according to statistics of the 2018 Fall Season [2];

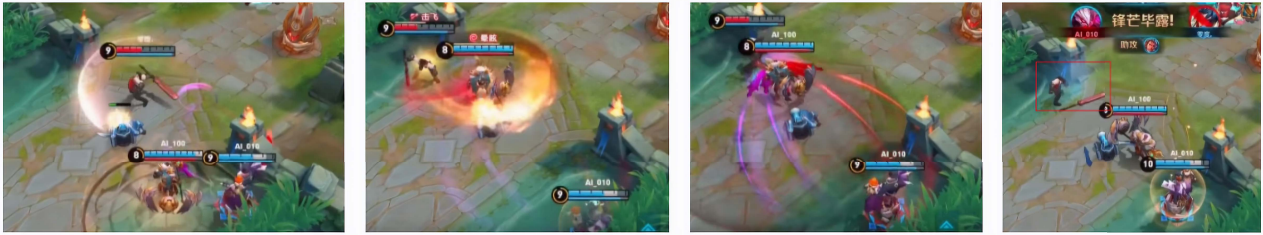

Fig. 8. Multiagent collaboration. Two AI heroes, AI-1 and AI-2, are in the blue health bar; the enemy hero in the red bar. (a) AI-1 protects AI-2 with a shield buff to guarantee safety. (b) and (c) AI-1 forces the enemy to change move direction with its skill, creating space for AI-2 to lock the target with its ultimate skill. (d) AI-2's skill hits, and the enemy gets killed (subfigures a, b, c, and d, from left to right).

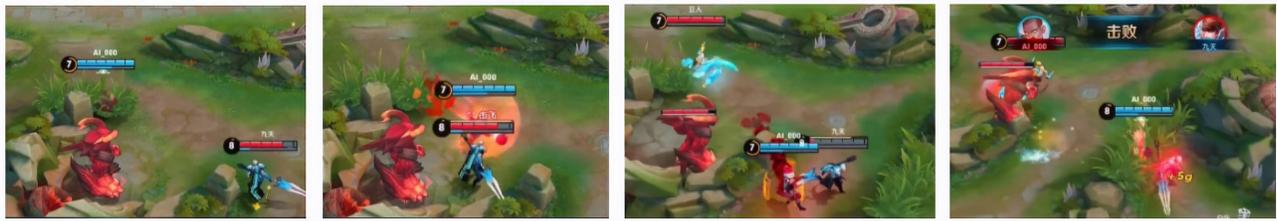

Fig. 9. Ambush with a critical strike. The AI hero is in the blue health bar; the enemy hero in the red bar. (a) AI hides in the bush, invisible to the enemy. (b) and (c) AI assaults with a deadly skill combo. (d) enemy hero killed (subfigures a, b, c, and d, from left to right).

TABLE III
MATCH BETWEEN AI AND TOP HUMAN TEAM

| Date | Red | Blue | Winner | Kill (AI:Human) |
|---|---|---|---|---|
| 07/11/18 | JueWu-SL | Human | JueWu-SL | 17 : 11 |
| 07/11/18 | Human | JueWu-SL | JueWu-SL | 34 : 19 |
| 08/11/18 | JueWu-SL | Human | JueWu-SL | 25 : 22 |
| 08/11/18 | Human | JueWu-SL | JueWu-SL | 28 : 18 |
| 09/11/18 | JueWu-SL | Human | Human | **22 : 23** |
| 09/11/18 | Human | JueWu-SL | Human | **32 : 37** |
| 15/11/18 | JueWu-SL | Human | JueWu-SL | 27 : 23 |
| 15/11/18 | Human | JueWu-SL | JueWu-SL | 27 : 24 |
| 16/11/18 | JueWu-SL | Human | JueWu-SL | 30 : 20 |
| 16/11/18 | Human | JueWu-SL | JueWu-SL | 26 : 14 |

note that the proportion of players in each game level may slightly vary from season to season).

The match results are listed in Table III. AI dominantly won the first five matches, demonstrating a strong level of micromanagement and team collaboration. However, it was defeated consecutively in Game 5 and Game 6. In both games, the human team adopted an attempted opening strategy by invading the Jungle of the AI team, followed by turret pushing in the middle. The AI could not handle this Jungle invading very well in the opening. As a result, the human team quickly gained an advantage in gold and experience. Then, the human players' huddle, i.e., stay together to protect each other. Finally, the human team was able to win the game even though the team kill was not dominant, as shown in Table III. To tackle this problem, we make the following adjustments: 1) divide the Jungle region into two parts, i.e., home jungle and opponent jungle; 2) extract data containing team fights happen in the home jungle using our scene-based sampling method; and 3) fine-tune the trained model with these data. After these adjustments, we continued the match. It was observed that AI learned how to handle the aforementioned bad case and won the next four games in a row.

In the KPL Finals of 2018 (KPL is short for King Pro League, which is an annual Honor of Kings Esports tournament. KPL to Honor of Kings is similar to TI (The International) to Dota, we invited a semiprofessional team that consists of five celebrated human players in the community (two former professionals and three well-known Esports players) to play against JueWu-SL. The match was broadcast live and worldwide on December 22, 2018. The match lasted for 12 min 59 s. AI won the match by a team pushing from the middle lane. The total team kill was 12-to-12. The number of turrets destroyed and the team gold between AI and human were 6-to-4 and 38.9k-to-35.6k, respectively.

In these matches, we observe that AI learned many high-level human-like game tactics and can execute these tactics well. Here, we analyze a few game episodes between AI and the semiprofessional human team to illustrate the tactics that AI has learned. In Fig. 8, one example of AI's team collaboration is shown. We can observe in detail how two AI heroes collaborate using move and skills to kill a hero controlled by a human. In Fig. 9, we show a successful ambush of AI. The AI hero first hides in the bush, waiting for the enemy to come. Once the target appears, AI precisely releases a combo of skills, creating vast damage to the enemy hero. Then, AI finishes a kill with a pursuit.

Some matches played by our AI are publicly available at https://sourl.cn/kjmPk7. Through these matches, one can see how it actually performs.

*3) Playing Strength With Human Teammates:* For single AI bot ability evaluation, we test the case that one AI

TABLE IV
ABLATION STUDY

| Ablation | Win rate VS full [CI] |
| --- | --- |
| w/o scene-based sampling | 0.17 [0.10; 0.26] |
| w/o macro-strategy | 0.31 [0.22; 0.41] |
| w/o local image feature | 0.26 [0.18; 0.36] |
| w/o global image feature | 0.35 [0.26; 0.45] |
| w/o vector feature | 0.21 [0.13; 0.30] |

model cooperates with four King level human players to fight against a human team of five players of the same level. The opponents and teammates of AI are different groups of persons for different matches. According to our tests, our AI collaborates very well with high-level human teammates since it is trained using preprocessed human data. Taking the model of DiaoChan (a hero in Honor of Kings) as an example, it achieves a high win rate (70%) over 40 matches. Its damage to enemy heroes constitutes more than 30% of the total team damage, demonstrating that AI masters the game even when playing with a human. Similar performance can be observed from other AI heroes.

*4) Component Analysis:* We conduct ablation studies to analyze the effectiveness of each component in our method, including multimodal features, sampling methods, and macrointent labels. The results of 100 matches for each ablation are given in Table IV. We observe that all these components significantly contribute to the final version of our AI program. Specifically, the vector features and scene-based sampling are the most important. The win rate without these two components against the full version drops to 17% and 21%, respectively. We see that AI becomes weaker without learning macro-strategy as the auxiliary task (31% win rate), which demonstrates the usefulness of our modeling of player's intents during the game. In fact, these components are not proposed at once and are added incrementally with verified effectiveness, during the process of developing the AI program.

*C. Discussion*

We now discuss the underlying mechanism of the proposed method and some lessons learned.

Essentially, our AI is the collection of intelligence from top human players considered in the dataset. Under a supervised setting, AI is trained to synthesize an averaged strategy from human gameplays. In particular, the proposed AI model captures generalizable representations of both macro-strategy and micromanagement. Its macro-strategy component serves as a coordination signal for AI heroes and is implemented as an auxiliary task to its micromanagement component, which outputs the actions of each single AI. The multimodal features, including local and global image-like features and vector features, provide an informative description of the game state. Furthermore, from a huge amount of historical data that have thorough coverage of the game map and the heroes, we only choose well-performed games of top players (see Section V) rather than using them directly, and we sample scenes and actions taken within a scene (see Section IV-F). As mentioned earlier, the raw data utilization is only one-twentieth after processing. With such modeling and processing, AI is equipped with human intent and knowledge to cope with the varying game situations. Consequently, although the raw dataset for training is from the top 1% human players, the resulting AI outperforms the top 1% players in real games.

Recall that the aim of this article is to provide answers to an unexplored yet important Game AI problem: whether and how SL, which is a widely used Game AI method, works for MOBA games? Furthermore, what are the key techniques to build competitive MOBA Game AI agents that imitate human data? While exploring the performance upper bounds of both RL and SL for MOBA Game AI is beyond the scope of this paper, we qualitatively compare these two techniques to provide guidance on which to choose. Compared with RL, SL has an advantage in several aspects. 1) *Fast Manufacturing*. The exploration of RL in MOBA can be extremely expensive due to its giant complexity. A recent RL-based Dota AI [4] was trained continuously trained for as long as ten months using a massive amount of computing resources, i.e., thousands of GPUs. By comparison, SL can be trained incredibly faster to achieve the top player level. As mentioned earlier, given top human data, our AI is trained for only one to days using affordable resources, i.e., 16 GPUs. 2) *Ability to Scale*. For SL, AI can be trained hero by hero, with each hero using corresponding human data. Hence, methodologically, it is not hard to cover all the heroes in a MOBA game. For RL, a typical way of generating training data is through self-play [3], [4], which requires determining a pool of heroes before training. Existing RL systems have failed to scale up the number of heroes in the pool. For example, OpenAI Five only supported 17 heroes for Dota 2, the AI ability decreased when adding more heroes [4]. 3) *Application*. Naturally, the strategy of AI using human data tends to be human-like, which can easily be applied to real-world applications. In contrast, the strategy played by RL-based AI has obvious distinctions to that of human players [4]. On the other hand, one drawback of SL is the requirement of high-quality human data. Furthermore, SL's upper limit is theoretically lower than the RL counterpart, with a precondition of having massive resources and time for RL training. It becomes a tradeoff between computing resources and performance when choosing between RL and SL.

VI. CONCLUSION AND FUTURE WORK

This article introduces the design process of the AI agent, JueWu-SL, for playing standard MOBA 5v5 games that are considered as one of the most complex game types at present. We formulate the action-taken process of a MOBA agent as a hierarchical multiclass classification problem. We develop a supervised model to solve the problem, leveraging neural networks on top of both intent and action labels to simultaneously capture the intrinsics of macro-strategy and micromanagement of MOBA from experienced human data. We show that, for the first time, supervised AI agents can achieve top human-level performance in MOBA games. Although we use Honor of Kings as a case study, the problem formulation, the proposed techniques, and the lessons learned are also applicable to other MOBA games.

We expect JueWu-SL to encourage further methods using SL or RL to build AI agents for MOBA. For example, our network structure can be used as the policy network in RL. Also, our model can be easily adapted to provide suitable initializations in an RL setting and can also serve as the opponent for guiding RL's policy training. As an ongoing step, we are combining SL and RL to further improve our AI's ability. We are also exploring real-world applications that can benefit from our Game AI method.

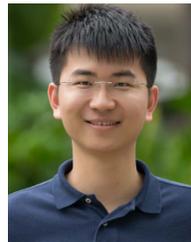

**Deheng Ye** finished his Ph.D. from the School of Computer Science and Engineering, Nanyang Technological University, Singapore, in 2016.

He is a Researcher and Team Manager with the Tencent AI Lab, Shenzhen, China, where he is currently in charge of the JueWu AI Project. He is broadly interested in applied machine learning, reinforcement learning, and software engineering.

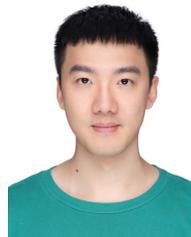

**Guibin Chen** received the M.S. degree *(cum laude)* from the Eindhoven University of Technology, Eindhoven, The Netherlands, in 2015.

He is currently a Researcher with the Tencent AI Lab, Shenzhen, China. Before joining Tencent, he was a Research Associate at Nanyang Technological University, Singapore. He is broadly interested in game intelligence and reinforcement learning.

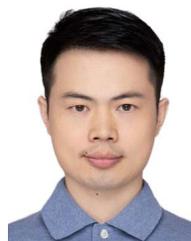

**Peilin Zhao** received the Ph.D. degree from Nanyang Technological University (NTU), Singapore.

He is currently a Researcher with the Tencent AI Lab, Shenzhen, China. Previously, he has worked at Rutgers University, Institute for Infocomm Research (I2R), and Ant Group. He has published over 100 papers in top venues, including JMLR, ICML, KDD, and so on. He has been invited as a PC member, reviewer, or editor for many conferences and journals, e.g., ICML, JMLR, and so on. His research interests include online learning, recommendation systems, automatic machine learning, deep graph learning, reinforcement learning, and so on.

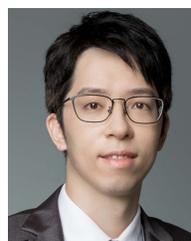

**Fuhao Qiu** received the B.S. and M.S. degrees from the South China University of Technology, Guangzhou, China, in 2012 and 2015, respectively.

He is currently a Researcher with the Tencent AI Lab, Shenzhen, China. His research interest is reinforcement learning, specifically exploration, communication, and multiagent systems in games.

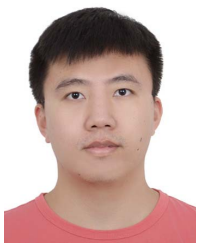

**Bo Yuan** received the B.S. and M.S. degrees from the Harbin Institute of Technology, Harbin, China, in 2012 and 2015, respectively.

He is currently a Researcher with the Tencent AI Lab, Shenzhen, China. His research interests include Game AI and data analysis.

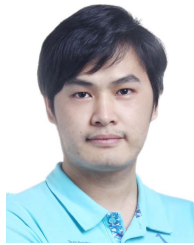

**Zhenjie Lian** received the B.S. degree in computer science from South China Agricultural University, Guangzhou, China, in 2008.

He is a Researcher with the Tencent AI Lab, Shenzhen, China. He is interested in system building, search engine optimization, and AI for video games.

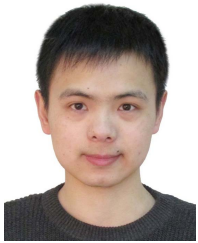

**Wen Zhang** received the M.S. degree from Zhejiang University, Hangzhou, China, in 2018.

He is a Researcher with the Tencent AI Lab, Shenzhen, China. His major interests include reinforcement learning, data mining, and neuroscience.

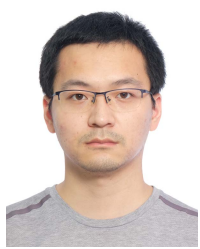

**Bei Shi** received the Ph.D. degree in computer science from The Chinese University of Hong Kong, Hong Kong, in 2018.

He is currently a Researcher with the Tencent AI Lab, Shenzhen, China. His research mainly focuses on reinforcement learning, natural language processing, and text mining.

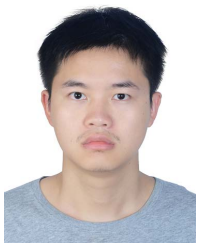

**Sheng Chen** received the B.S. and M.S. degrees from the University of Science and Technology of China, Hefei, China, in 2015 and 2018, respectively.

He is a Researcher with the Tencent AI Lab, Shenzhen, China. His research interests include Game AI and applied machine learning.

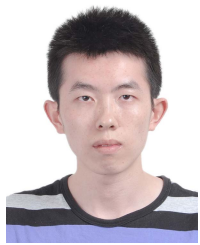

**Liang Wang** received the M.S. degree from the Huazhong University of Science and Technology, Wuhan, China, in 2011.

He is currently a Researcher and Team Manager with the Tencent AI Lab, Shenzhen, China. He is broadly interested in big data analytics and reinforcement learning.

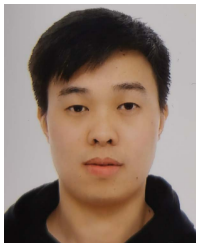

**Mingfei Sun** received the Ph.D. degree in computer science and engineering from the Hong Kong University of Science and Technology, Hong Kong, in 2020.

He is currently a Post-Doctoral Researcher at Oxford University, Oxford, U.K. (jointly with Microsoft Research Cambridge). Previously, he was an Intern with the Tencent AI Lab, Shenzhen, China, where this work was done.

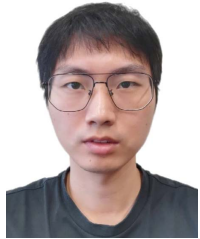

**Tengfei Shi** received the B.S. degree from Xiamen University, Xiamen, China, in 2009, and the M.S. degree from the Beijing University of Posts and Telecommunication, Beijing, China, in 2012.

He is an Engineer and Team Manager with the Tencent AI Lab, Shenzhen, China. His research interests are Game AI and large-scale learning systems.

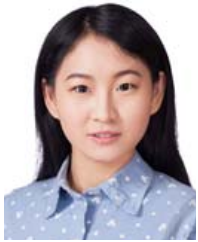

**Xiaoqian Li** received the B.S. and M.S. degrees from the Harbin Institute of Technology, Harbin, China, in 2014 and 2016, respectively.

She is currently a Researcher with the Tencent AI Lab, Shenzhen, China. Her research interests include Game AI and applied machine learning.

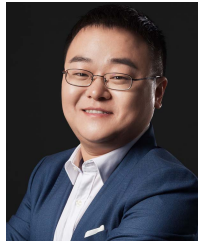

**Qiang Fu** received the B.S. and M.S. degrees from the University of Science and Technology of China, Hefei, China, in 2006 and 2009, respectively.

He is the Director of the Game AI Center, Tencent AI Lab, Shenzhen, China. He has been dedicated to machine learning, data mining, and information retrieval for over a decade. His current research focus is game intelligence and its applications, leveraging deep learning, domain data analysis, reinforcement learning, and game theory.

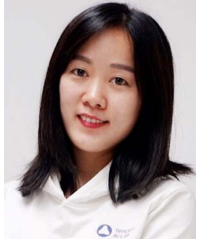

**Siqin Li** received the B.S. and M.S. degrees from the Harbin Institute of Technology, Harbin, China, in 2013 and 2015, respectively.

She is currently a Researcher with the Tencent AI Lab, Shenzhen, China. Her research interests include Game AI and machine learning.

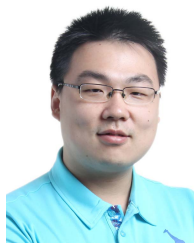

**Wei Yang** received the M.S. degree from the Huazhong University of Science and Technology, Wuhan, China, in 2007.

He is currently the General Manager of the Tencent AI Lab, Shenzhen, China. He has pioneered many influential projects in Tencent in a wide range of domains, covering Game AI, Medical AI, search, data mining, large-scale learning systems, and so on.

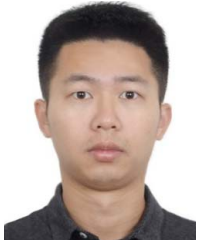

**Jing Liang** received the B.S. and M.S. degrees from the Beijing University of Posts and Telecommunication, Beijing, China, in 2011 and 2014, respectively.

He is a Researcher with the Tencent AI Lab, Shenzhen, China. His research interests include machine learning and pattern recognition.

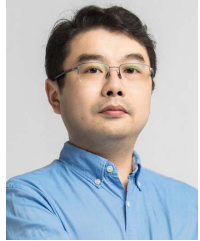

**Lanxiao Huang** received the B.S. and M.S. degrees from the University of Electronic Science and Technology of China, Chengdu, China, in 2005 and 2008, respectively.

He is the General Manager of the Tencent Timi L1 Studio and the Executive Producer of Honor of Kings. He focuses on game design and applications of AI technology in video games.